\documentclass[10pt,twocolumn,letterpaper]{article}

\usepackage[pagenumbers]{cvpr}  %

\usepackage{cuted} %

\usepackage{currfile} %

\usepackage{caption} %

\usepackage{booktabs}
\usepackage{colortbl}
\usepackage{multirow}
\usepackage{xspace}
\usepackage{amssymb}
\usepackage[skip=2pt]{subcaption}
\usepackage{enumitem}

\newcommand{\ours}{\textsc{AeroRSI}\xspace}
\newcommand{\afa}{AFA\xspace}
\newcommand{\cvr}{CVR\xspace}
\newcommand{\rrl}{RRL\xspace}

\newcommand{\up}{$\uparrow$}
\newcommand{\down}{$\downarrow$}
\newcommand{\bd}[1]{\textbf{#1}}
\definecolor{oursrow}{gray}{0.93}

\usepackage{algorithm}
\usepackage[noend]{algpseudocode}

\definecolor{cvprblue}{rgb}{0.21,0.49,0.74}
\usepackage[pagebackref,breaklinks,colorlinks,allcolors=cvprblue]{hyperref}

\title{AeroRSI: Recursive Self-Improvement for Aerial Vision-Language Navigation}

\makeatletter
\def\@maketitle{%
  \newpage\null\vskip .2in
  \begin{center}
    {\Large\bf \@title\par}
    \vspace*{12pt}
    {\large\lineskip .4em \@author\par}
    \vspace*{4pt}
  \end{center}}
\makeatother

\author{%
Shiwei Wu$^{1}$\hspace{0.85em}Yuwen Pan$^{1}$\hspace{0.85em}Zhi Li$^{1}$\hspace{0.85em}Yaowen Li$^{2}$\hspace{0.85em}%
Yan Zhou$^{3}$\hspace{0.85em}Xuemiao Xu$^{4}$\hspace{0.85em}Yu Liu$^{2,\dagger}$\\[0.25em]
{\small $^{1}$SIGS, Tsinghua University\hspace{1.2em}$^{2}$Department of Electronic Engineering, Tsinghua University}\\[0.05em]
{\small $^{3}$Foshan University\hspace{1.2em}$^{4}$South China University of Technology\hspace{1.2em}$^{\dagger}$Corresponding author}\\[0.15em]
{\small Project code: \url{https://github.com/starsholic/AeroRSI}}%
}

\newskip\aerorsiFloatsep \newskip\aerorsiTextFloatsep \newskip\aerorsiIntextsep
\newskip\aerorsiDblFloatsep \newskip\aerorsiDblTextFloatsep
\AtBeginDocument{\aerorsiFloatsep=\floatsep \aerorsiTextFloatsep=\textfloatsep \aerorsiIntextsep=\intextsep
  \aerorsiDblFloatsep=\dblfloatsep \aerorsiDblTextFloatsep=\dbltextfloatsep}

\begin{document}
\twocolumn[{%
\renewcommand\twocolumn[1][]{#1}%
\maketitle
\vspace{-1.6em}%
\begin{center}
\includegraphics[width=\textwidth]{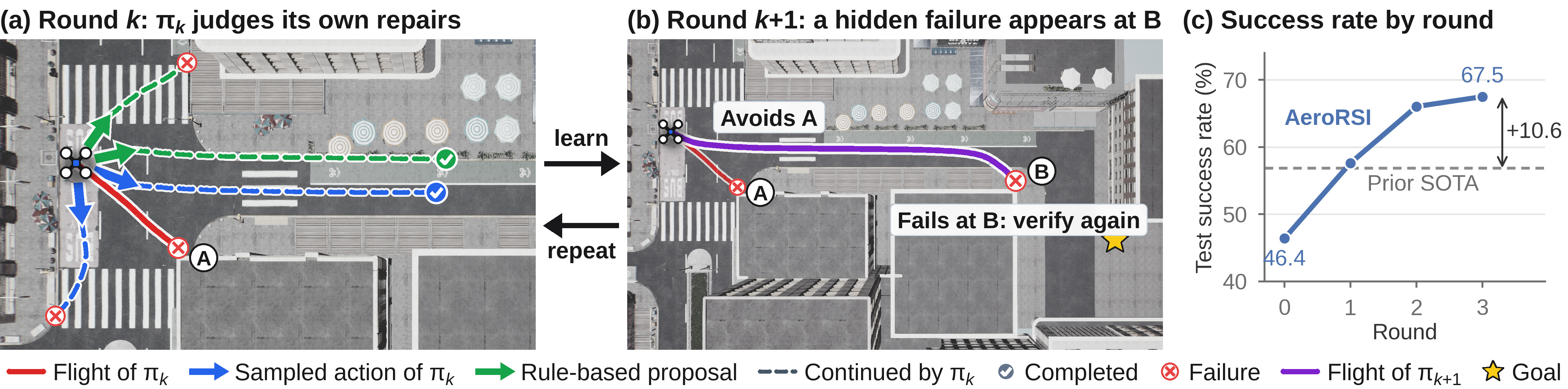}
\captionof{figure}{\textbf{Recursive self-improvement with self-verified repairs.} (a)~In round $k$, $\pi_k$ fails at A and judges its own candidate repairs. From the state before A, each candidate from its sampled actions or rule-based proposals is continued by $\pi_k$, and only those it completes are kept. (b)~After learning these repairs, $\pi_{k+1}$ avoids A and reaches B, a failure concealed by the earlier one at A, where the next round verifies new repairs. (c)~Test success rate on TravelUAV; the dashed line is the previous state of the art~\cite{xu2026aerodpo} from the same supervised policy.}
\label{fig:teaser}
\end{center}
\vspace{0.4em}%
}]
\begin{abstract}
Vision-language-action (VLA) models enable unmanned aerial vehicles (UAVs) to follow natural-language instructions, yet still fail frequently in closed-loop flight.
Correcting these failures requires repairs, \ie, actions that avoid them at the states the policy itself reaches, which human pilots cannot label at scale and rule-based planners cannot supply reliably.
In this paper, we present \ours, a recursive self-improvement framework in which an aerial navigation policy iteratively learns from its own failures.
Our key insight is that a failing policy should judge its own repairs by flying on from them.
Judged in this way, its sampled actions and even rule-based proposals that hurt it when imitated directly become reliable supervision.
Moreover, an episode terminates at its first failure, concealing subsequent failures until the earlier ones are repaired.
\ours thus renews its supervision in every round, anchoring the failures of the current policy, verifying repairs by its continuation, and distilling them while keeping its behavior elsewhere.
On TravelUAV, three rounds raise the success rate of a supervised 2B-parameter policy from 46.4\% to 67.5\% and reduce collisions by half, surpassing the previous state of the art by 10.6 points from the same initialization without extra human labels or inference cost.
\par\vfill
\end{abstract}

\section{Introduction}
\label{sec:intro}

Unmanned aerial vehicles (UAVs) that follow natural-language instructions could inspect infrastructure, search for people, and deliver goods without a human pilot.
Toward this goal, recent vision-language-action (VLA) models for aerial vision-language navigation (VLN)~\cite{liu2023aerialvln,wang2025travelUAV,gao2025openfly,lee2025citynav,wang2025uavflow,jiang2026satnav} directly map visual observations and instructions to flight actions~\cite{zhang2026navfom,xu2026aerialvla,xu2026aerodpo,sun2026autofly,cai2025flightgpt}, mirroring their success in robotic manipulation~\cite{brohan2023rt1,brohan2023rt2,octo2024octo,kim2024openvla,black2025pi0}.
Despite this progress, existing aerial agents~\cite{zhang2026navfom,xu2026aerodpo,jiang2025longfly,xu2026aerialvla,wang2026refinefly} still fail frequently in closed-loop flight, often due to collisions with obstacles.
Since a single wrong action can terminate an entire flight, such failures severely limit their reliability.

These failures stem from a well-known mismatch in imitation learning~\cite{pomerleau1988alvinn,ross2010efficient,ross2011dagger}, as a policy trained on demonstrations reaches states that the demonstrations never cover and accumulates errors along the flight.
Correcting them requires \emph{repairs}, namely actions that avoid a failure at the states the policy itself reaches, and existing approaches differ mainly in who judges such an action.
Interactive imitation learning~\cite{ross2011dagger,kelly2019hgdagger,laskey2017dart,hoque2021thriftydagger} queries an expert for corrective actions, but human annotation is extremely expensive at scale, and rule-based planners~\cite{khatib1986apf,zhou2021ego} ignore the language instruction.
Automatic correction methods~\cite{xu2026aerodpo,yu2026correctnav,xiao2025pld} avoid human labeling by deriving corrections from collision rollbacks, reference paths, or auxiliary teacher policies, yet they do not consider whether the policy can actually execute these corrections.
Reinforcement learning~\cite{wang2026refinefly,ju2026xfly,zhao2026worldvln,ma2026semantic} instead relies on episode-level rewards, which indicate whether a flight fails but not which action should have been taken.
As a result, annotation-free, action-level supervision that the policy itself can execute remains missing.

Recursive self-improvement offers a promising route toward such supervision.
It has recently achieved remarkable success in language models~\cite{zelikman2022star,gulcehre2023rest,singh2024beyond,yuan2024selfrewarding,kumar2025training,deepseek2025r1}, where a model learns from its own verified outputs, and has begun to extend to robotics and navigation~\cite{li2026rsinav,zhang2026recovery,wang2026adarsi,lang2026embodirsi,he2026criticality,liu2026mobiagent,xiao2025pld}.
However, closed-loop flight poses three challenges that answer-level self-improvement leaves open.
First, an answer can be verified on its own, whereas the value of a flight action depends on the subsequent flight of the policy, which calls for step-level values as in process supervision~\cite{wang2024mathshepherd,kazemnejad2025vineppo}.
Second, a language model learns from a fixed set of problems, whereas a flight terminates at its first failure, so later failures remain hidden until earlier ones are repaired.
Third, self-improvement in language models mainly reinforces answers that the model can already produce~\cite{huang2025sharpening,song2025mind}, whereas a failing flight policy often cannot sample the action it needs, even though simple physical priors, such as climbing or detouring around obstacles, can readily propose one, as rationalization does in STaR~\cite{zelikman2022star}.

These challenges lead to our key insight that \emph{a failing policy should judge its own repairs}.
By continuing its flight after each candidate action, the current policy estimates the value of the candidate under its own behavior, which is what approximate policy iteration requires~\cite{kakade2002approximately,fern2006api,scherrer2012ampi}.
This self-verification turns weak sources into reliable supervision.
Actions sampled from the policy often contain a repair, and rule-based proposals that degrade the policy under direct imitation become its most effective supervision once verified, whereas checking only the next few actions often favors candidates that fail later (Fig.~\ref{fig:teaser}(a)).
It also reveals how failures evolve.
Verified repairs remain valid for subsequent policies, while new failures emerge elsewhere on the same routes (Fig.~\ref{fig:teaser}(b)), so supervision must be renewed in every round.
Moreover, actions learned from rule-based proposals in one round appear among its sampled actions in the next, indicating that the loop expands what the policy can do rather than reinforcing existing behavior.

Based on these insights, we present \ours, a recursive self-improvement framework that repeats three steps in every round.
First, adaptive failure anchoring (\afa) collects the states leading to each failure of the current policy at multiple time scales, so that supervision follows the failures as they are revealed.
Starting from these states, continuation-verified repair (\cvr) builds candidates from actions sampled from the policy and from rule-based proposals, and accepts a candidate only if the original action fails again from the same state or falls clearly behind the best candidate, and the policy completes the candidate again in an independent re-flight.
Finally, retentive repair learning (\rrl) distills these verified repairs into the policy while keeping its behavior on states that need no repair.

Extensive experiments on the TravelUAV benchmark~\cite{wang2025travelUAV} demonstrate the effectiveness of \ours.
Starting from a supervised policy, \ours improves the success rate from 46.4\% to 67.5\% in three rounds and reduces collisions by half (Fig.~\ref{fig:teaser}(c)), outperforming the previous state-of-the-art method~\cite{xu2026aerodpo}, which is trained from the same policy, by 10.6 points.
When built upon the stronger policy of this method, \ours further improves the success rate to 69.9\%, establishing a new state of the art without additional human annotation or inference cost.

In summary, our contributions are as follows.
\begin{itemize}[leftmargin=*,topsep=2pt,itemsep=1pt]
\item We identify three challenges of self-improvement in closed-loop flight and show that letting a failing policy judge its own repairs turns its sampled actions and even rule-based proposals into reliable supervision.
\item We reveal that verified repairs remain valid while new failures emerge elsewhere on the routes, and that actions learned from rule-based proposals enter its sampled actions, which together make recursion essential.
\item We propose \ours, which anchors the failures of the current policy, verifies candidate repairs by its own continuation, and distills them while keeping its behavior elsewhere, achieving state-of-the-art results on TravelUAV with no extra annotation or inference cost.
\end{itemize}

\section{Related Work}
\label{sec:related}

We review three lines of work most related to ours, namely aerial vision-language navigation, learning from corrections, and recursive self-improvement.

\noindent\textbf{Aerial Vision-Language Navigation.}
Vision-language navigation was first studied for ground agents~\cite{anderson2018vln,fried2018speaker,chen2022think,wang2023scaling,krantz2020beyond,an2024etpnav,shah2022lmnav} and later extended to UAVs by AerialVLN~\cite{liu2023aerialvln}, TravelUAV~\cite{wang2025travelUAV}, OpenFly~\cite{gao2025openfly}, and subsequent benchmarks~\cite{lee2025citynav,wang2025uavflow,jiang2026satnav}.
Recent methods improve cross-embodiment pretraining~\cite{zhang2026navfom}, long-horizon context modeling~\cite{jiang2025longfly}, robustness to error accumulation~\cite{tang2026neurokalman}, future prediction~\cite{wang2026forefly,deng2026dreamfly,zheng2026worldfly,liu2026imagineuav,liu2026airforesight}, and spatial reasoning~\cite{gao2024exploring,cai2025flightgpt,wu2025aeroduo}.
AerialVLA~\cite{xu2026aerialvla} and AeroDPO~\cite{xu2026aerodpo} show that lightweight vision-language-action policies can match much larger models, and post-training further improves them through preference optimization on synthesized corrections~\cite{xu2026aerodpo} or online reinforcement learning~\cite{wang2026refinefly,ju2026xfly,zhao2026worldvln,ma2026semantic}.
These efforts mostly improve the model architecture or learn from synthesized corrections and episode-level rewards, whereas \ours learns from repairs that the policy verifies itself, collected anew from its failures in every round, and is complementary to these architectures.

\noindent\textbf{Learning from Corrections.}
Interactive imitation learning asks an expert to label the states that the learner itself encounters, as in DAgger~\cite{ross2011dagger} and its variants~\cite{kelly2019hgdagger,laskey2017dart,hoque2021thriftydagger}, while AggreVaTe~\cite{ross2014aggrevate} ranks actions by the cost-to-go of the expert.
Navigation policies have also been learned from expert corrections~\cite{ross2013learning} from crashes and disengagements~\cite{gandhi2017learning,kahn2021badgr,kahn2021land}, and in simulation~\cite{sadeghi2017cad2rl,loquercio2021learning,kaufmann2023champion}.
To avoid human labeling, recent methods synthesize corrections by collision rollbacks~\cite{xu2026aerodpo}, reference paths~\cite{yu2026correctnav}, or auxiliary residual policies~\cite{xiao2025pld}, and learn from them with supervised, preference, or reinforcement objectives~\cite{ouyang2022training,rafailov2023dpo,hong2024orpo,shao2024deepseekmath}, although preference updates may shift probability to unintended actions~\cite{razin2025unintentional,tajwar2024preference,pang2024iterative}.
Since these corrections are judged by an expert, a heuristic, a reference path, or another policy, none of them accounts for whether the policy itself can execute them.
\cvr instead judges each candidate by how the current policy, not an expert, continues the flight after it, and \rrl imitates the verified best candidate rather than contrasting it with the failed action.

\noindent\textbf{Recursive Self-Improvement.}
Training a model on its own verified outputs underlies expert iteration~\cite{anthony2017thinking,silver2017mastering,silver2018general}, self-training~\cite{xie2020self}, STaR~\cite{zelikman2022star}, and ReST~\cite{gulcehre2023rest,singh2024beyond}, and repeated rounds of it have substantially improved the reasoning of language models~\cite{kumar2025training,deepseek2025r1,yuan2024selfrewarding,lightman2024lets}, although recent analyses find that it mainly reinforces outputs that the model can already produce~\cite{huang2025sharpening,song2025mind}.
Weak supervisors can still teach stronger models~\cite{burns2024weak}, and autocurricula create new challenges as learning proceeds~\cite{baker2020emergent}.
In robotics and navigation, agents likewise learn from data that they generate themselves~\cite{bousmalis2023robocat,zhou2024soar,ghasemipour2025selfimproving}, obtain corrections from residual policies~\cite{xiao2025pld}, build on coding or multimodal agents~\cite{wang2026adarsi,lang2026embodirsi,liu2026mobiagent}, direct data collection with learned models~\cite{he2026criticality}, or retrain on their own trajectories and reasoning~\cite{li2025sid,lin2025evolvenav,li2026rsinav}, with general formulations in~\cite{zhou2026who,tang2026gai}.
Closest to ours, CorrectNav~\cite{yu2026correctnav} repeatedly retrains a policy on corrections of its errors toward reference paths, and local recovery supervision~\cite{zhang2026recovery} checks whether a frozen student can continue from a teacher's recoveries.
Rollout-based policy iteration~\cite{fern2006api,scherrer2012ampi} and learning to search~\cite{daume2009searn,chang2015lols,sun2017aggrevated} also judge actions by a continuation, but roll out every action of a small set or follow a reference policy.
\ours brings this judgment to the failures of a VLA policy with nearly a million actions, verifying candidates from its samples and physical priors by Monte-Carlo continuations from restored states~\cite{tesauro1996online,bertsekas1997rollout,ecoffet2021first}, without reference paths or a teacher, and \rrl retains its behavior elsewhere, as in continual learning~\cite{li2018learning,kirkpatrick2017overcoming,rolnick2019experience}.
Our ablations compare these choices with such combinations (Sec.~\ref{sec:exp_ablation}): unverified rule-based corrections help little or even hurt, checking only the first actions leaves more collisions, and reusing one round of supervision falls short of anchoring anew in every round.

\section{Method}
\label{sec:method}

\begin{figure*}[t]
\centering
\includegraphics[width=\linewidth,trim=7.2pt 0pt 7.2pt 0pt,clip]{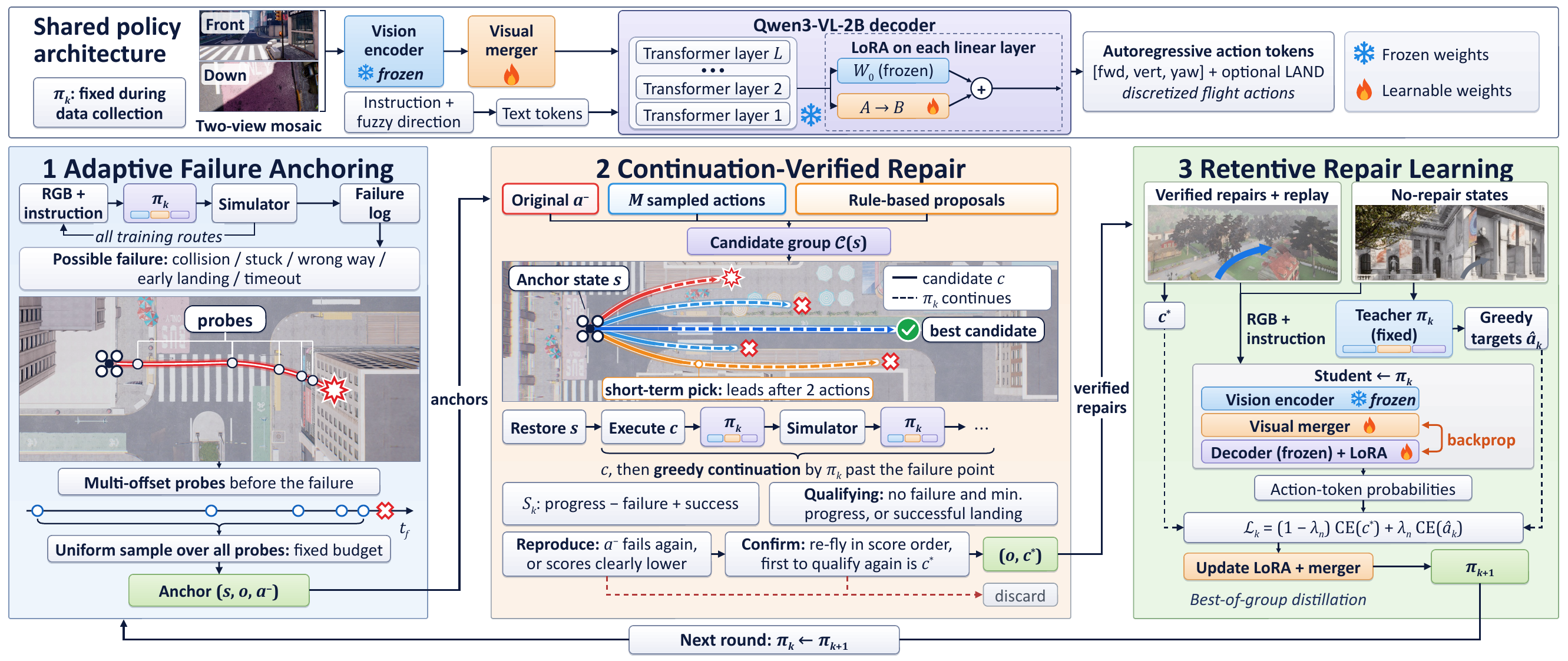}
\caption{\textbf{Overview of \ours.} The top row shows the policy $\pi_k$, of which only the visual merger and the LoRA are trained. In each round, (1)~\afa samples anchors from probes before every failure of $\pi_k$; (2)~\cvr restores each anchor, lets $\pi_k$ continue the executed action $a^{-}$, its sampled actions, and rule-based proposals, and accepts as the repair $c^{\star}$ the first candidate that qualifies again in a re-flight; and (3)~\rrl distills the verified repairs into a student copy of $\pi_k$, keeping the outputs of $\pi_k$ on states that need no repair, to obtain $\pi_{k+1}$.}
\label{fig:framework}
\end{figure*}

In this section, we first formulate recursive self-improvement for aerial navigation as approximate policy iteration and then present the three modules of \ours that implement it in every round.

\subsection{Problem Formulation}
\label{sec:method_problem}

\noindent\textbf{Task.}
At each step $t$, the UAV receives an observation $o_t$ consisting of its front and downward RGB views and the instruction, together with a coarse direction to the target as in the fuzzy direction guidance of~\cite{xu2026aerodpo}.
The policy $\pi$ outputs an action $a_t$ that discretizes the forward, vertical, and yaw motion, optionally with a landing action.
An episode succeeds if the UAV lands near the described target, and it terminates at the first failure, namely a collision, getting stuck, moving away from the target, landing prematurely, or exceeding the step limit.
During training, the simulator can save and restore the full state $s_t$ of the UAV and its environment and provides depth maps and the target position.
\ours uses these privileges only to generate training data, and the deployed policy observes $o_t$ alone, with the architecture and inference cost of the initial policy $\pi_0$.

\noindent\textbf{Recursive Self-Improvement.}
Starting from $\pi_0$, each round $k$ generates supervision $\mathcal{D}_k$ from the experience of the current policy $\pi_k$ without human annotation and uses $\mathcal{D}_k$ to update it to $\pi_{k+1}$.
From the perspective of approximate policy iteration~\cite{kakade2002approximately}, an ideal round would relabel every state $s$ at which $\pi_k$ fails with the improved action
\begin{equation}
a^{\star}(s) = \arg\max_{a}\, Q^{\pi_k}(s, a),
\label{eq:improve}
\end{equation}
where $Q^{\pi_k}(s,a)$ is the expected return of executing $a$ in $s$ and following $\pi_k$ afterward.
Because $Q^{\pi_k}$ is defined by the subsequent flight of $\pi_k$ itself, it is the failing policy that should judge its own repairs, as in rollout-based policy iteration~\cite{fern2006api,scherrer2012ampi} and learning to search~\cite{daume2009searn,chang2015lols}.
Since the judge is the policy itself, each round also improves the next judge, which makes the loop recursive.
Two properties of closed-loop flight make this target hard to obtain.
The outcome of a continuation is stochastic, so a single rollout is an unreliable estimate of $Q^{\pi_k}$.
Moreover, as an episode terminates at its first failure, the states that need repair are revealed only as $\pi_k$ improves.
\ours therefore estimates $Q^{\pi_k}$ from verified continuations of the current policy and relabels the newly revealed failures in every round.

\subsection{Overview}
\label{sec:method_overview}

Fig.~\ref{fig:framework} illustrates one round of \ours.
The current policy $\pi_k$ first flies all training routes, and \afa selects anchor states leading to its failures.
\cvr then turns each anchor into a verified repair or discards it, and \rrl distills the repairs into $\pi_{k+1}$ while keeping the behavior of $\pi_k$ on states that need no repair, and the next round starts from the failures that $\pi_{k+1}$ reveals.

\subsection{Adaptive Failure Anchoring (\afa)}
\label{sec:method_afa}

Hand-designed anchors, such as the last action before a collision, cover only the failure types and time scales that a designer anticipates, whereas the failures of the policy change as it improves.
\afa therefore anchors every failure revealed by the current policy at multiple time scales, regardless of its type.

For each training episode of $\pi_k$ that ends in a failure, \afa places probes at exponentially spaced offsets before the end of the episode.
Near offsets capture immediate causes, such as an action into an obstacle, and far offsets capture earlier decisions, such as a turn into a dead end.
Each probe records the saved state $s$, the observation $o$, and the action $a^{-}$ that $\pi_k$ executed.
Since the probes far outnumber the anchors that can be verified, \afa draws a fixed budget of anchors uniformly from all probes of the round.
The anchors therefore follow the failure distribution of $\pi_k$ without hand-crafted rules, so that frequent failure types and time scales receive more anchors, and failures that the policy has overcome receive none.
Each anchor is finally assigned an evaluation horizon that grows with its offset and extends beyond the original failure point, so that a qualifying candidate must carry the flight past this point.

\subsection{Continuation-Verified Repair (\cvr)}
\label{sec:method_cvr}

\noindent\textbf{Candidate Group.}
For an anchor with state $s$, observation $o$, and original action $a^{-}$, \cvr builds the candidate group
\begin{equation}
\mathcal{C}(s) = \{a^{-}\} \cup \{a^{(i)} \sim \pi_k(\cdot \mid o)\}_{i=1}^{M} \cup \mathcal{P}(s),
\label{eq:group}
\end{equation}
which contains the original action, $M$ actions sampled from the current policy, and a set $\mathcal{P}(s)$ of rule-based proposals, namely a potential-field avoidance action~\cite{khatib1986apf} computed from the depth map and the target position, a variant of it that keeps the original forward motion, and a climb that keeps the original heading.
The sampled actions are ones that the policy can already produce but does not choose, and the rule-based proposals add avoidance actions that it rarely samples.
No candidate is trusted as a label before it passes verification.

\noindent\textbf{Continuation Score.}
To estimate $Q^{\pi_k}(s, c)$ for a candidate $c$, \cvr restores $s$, executes $c$, and lets the current policy continue greedily until the evaluation horizon of the anchor or the end of the episode.
The candidate receives the score
\begin{equation}
S_k(s, c) = \Delta d - \lambda_{f}\, \mathbb{I}[\text{fail}] + \lambda_{s}\, \mathbb{I}[\text{success}],
\label{eq:score}
\end{equation}
where $\Delta d$ is the progress toward the target, a failure denotes a collision, a premature landing, or an invalid output, and $\lambda_f$ and $\lambda_s$ weight failure and success.
Unlike a short-term criterion that inspects only the immediate outcome of $c$, $S_k$ judges a candidate by the subsequent flight of the policy itself and is thus a finite-horizon estimate of $Q^{\pi_k}$.

\noindent\textbf{Verified Acceptance.}
Since a single continuation is stochastic, \cvr accepts a repair only after two checks against chance outcomes.
A candidate qualifies if it reaches the end of the horizon without failure and with a minimum progress, or if it lands successfully, so that getting stuck or flying away counts as too little progress.
The reproduction check first replays the original action from $s$ and keeps the anchor only if this action fails within the horizon or the best qualifying candidate outscores it by a margin.
The confirmation check then re-flies the two best qualifying candidates from $s$ in order and accepts the first one, $c^{\star}$, that qualifies again.
Each accepted anchor yields a repair $(o, c^{\star})$, and the next round verifies the new failures of $\pi_{k+1}$.

\subsection{Retentive Repair Learning (\rrl)}
\label{sec:method_rrl}

A policy trained only on repairs tends to drift on states that need no repair, because it transfers the action statistics of the repairs, such as frequent turns, to the rest of the flight.
\rrl therefore updates the policy with three components that together limit this drift.

\noindent\textbf{Repair Supervision.}
\rrl maximizes the likelihood of the verified repair $c^{\star}$ of each anchor and, unlike preference optimization~\cite{rafailov2023dpo}, adds no penalty on the original action.
The original action is often only moderately worse than the repair and remains correct in many other states, so pushing it down explicitly risks degrading the flight elsewhere.

\noindent\textbf{Behavior Retention.}
\rrl adds a set $\mathcal{N}$ of states that need no repair, taken from successful training flights and covering early flight, the approach to the target, and the landing.
Their targets $\hat{a}_k(o)$ are the greedy outputs of the current policy $\pi_k$ rather than the demonstrations or an earlier policy, so that retention anchors the policy to its own behavior where no repair is needed without pulling it back to an outdated one.

\noindent\textbf{Repair Replay.}
From the second round on, \rrl replays the repairs of the previous round with a share $\rho$ of the repair weight as a safeguard against forgetting them.

Together, the update of round $k$ minimizes
\begin{equation}
\mathcal{L}_k = (1-\lambda_n)\, \mathbb{E}_{\mathcal{D}_k}\big[\mathrm{CE}(c^{\star})\big] + \lambda_n\, \mathbb{E}_{\mathcal{N}}\big[\mathrm{CE}(\hat{a}_k)\big],
\label{eq:loss}
\end{equation}
where CE is the cross-entropy of the action tokens given $o$, $\mathcal{D}_k$ draws a verified repair of round $k$ with probability $1-\rho$ and a repair of the previous round with probability $\rho$, and $\lambda_n$ is the share of retention.
The policy is initialized from $\pi_k$, and the trained policy becomes $\pi_{k+1}$.

\section{Experiments}
\label{sec:exp}
\setlength{\textfloatsep}{8pt plus 2pt minus 2pt}
\setlength{\dbltextfloatsep}{8pt plus 2pt minus 2pt}
\setlength{\floatsep}{8pt plus 2pt minus 2pt}
\setlength{\dblfloatsep}{8pt plus 2pt minus 2pt}
\setlength{\intextsep}{8pt plus 2pt minus 2pt}

In this section, we first describe the experimental setup and compare \ours with existing methods, then ablate its design, and finally present case studies of recorded flights.

\subsection{Experimental Setup}
\label{sec:exp_setup}

\begin{table}[t]
\centering
\footnotesize
\setlength{\tabcolsep}{3.5pt}
\caption{\textbf{Compared policies.} They share one 2B-parameter model with fuzzy direction guidance; the supervised policy (SFT) imitates the expert demonstrations of AerialVLA~\cite{xu2026aerialvla}, and the suffix of \ours names its initial policy.}
\label{tab:setup}
\begin{tabular}{@{}lll@{}}
\toprule
Policy & Initialized from & Trained on \\
\midrule
Supervised policy (SFT)~\cite{xu2026aerodpo} & Qwen3-VL-2B~\cite{bai2025qwen3vl} & Expert demonstrations \\
AeroDPO~\cite{xu2026aerodpo} & SFT & Collision corrections \\
\ours (SFT init.) & SFT & Verified repairs \\
\ours (DPO init.) & AeroDPO & Verified repairs \\
\bottomrule
\end{tabular}
\end{table}

\begin{table*}[t]
\centering
\footnotesize
\setlength{\tabcolsep}{4.0pt}
\caption{\textbf{Comparison with existing methods on the TravelUAV test splits.} L1: L1 assistant guidance of~\cite{wang2025travelUAV}; FD: fuzzy direction guidance of~\cite{xu2026aerodpo}. \ours (SFT init.) starts from the same supervised policy as AeroDPO, and \ours (DPO init.) from AeroDPO (Table~\ref{tab:setup}); both after three rounds. Overall: SR over all test routes, weighted by split sizes where only per-split results exist. Best results are in bold.}
\label{tab:main}
\begin{tabular}{llccccccccccccc}
\toprule
\multirow{2}{*}{Method} & \multirow{2}{*}{Guide} & \multicolumn{4}{c}{Seen} & \multicolumn{4}{c}{Unseen Object} & \multicolumn{4}{c}{Unseen Map} & Overall \\
\cmidrule(lr){3-6}\cmidrule(lr){7-10}\cmidrule(lr){11-14}\cmidrule(lr){15-15}
& & NE\down & SR\up & OSR\up & SPL\up & NE\down & SR\up & OSR\up & SPL\up & NE\down & SR\up & OSR\up & SPL\up & SR\up \\
\midrule
TravelUAV~\cite{wang2025travelUAV} & L1 & 106.28 & 16.10 & 44.26 & 14.30 & 118.11 & 22.42 & 46.90 & 20.51 & 138.80 & 4.18 & 20.77 & 3.84 & 13.62 \\
NavFoM~\cite{zhang2026navfom} & L1 & 93.05 & 29.17 & 49.24 & 25.03 & 108.04 & 29.83 & 47.99 & 27.20 & 125.10 & 6.30 & 18.95 & 5.68 & 22.02 \\
LongFly~\cite{jiang2025longfly} & L1 & 60.02 & 36.39 & 65.87 & 31.07 & 66.74 & 43.87 & 64.56 & 38.39 & 108.32 & 11.27 & 30.27 & 9.32 & 29.95 \\
NeuroKalman~\cite{tang2026neurokalman} & L1 & 71.56 & 25.86 & 58.73 & 22.43 & 71.01 & 32.48 & 60.82 & 28.50 & 100.32 & 8.34 & 34.15 & 7.12 & 21.66 \\
AerialVLA~\cite{xu2026aerialvla} & FD & 65.88 & 47.96 & 57.69 & 38.54 & 61.45 & 56.60 & 64.86 & 46.61 & 67.42 & 37.58 & 52.92 & 28.22 & 46.46 \\
RefineFly~\cite{wang2026refinefly} & FD & 54.96 & 56.33 & 65.40 & 45.98 & 53.42 & 59.72 & 68.31 & 51.34 & 58.88 & 42.97 & 60.09 & 31.83 & 52.78 \\
AeroDPO~\cite{xu2026aerodpo} & FD & 51.84 & 60.01 & 67.77 & 47.30 & 47.68 & 66.45 & 72.02 & 55.30 & 68.79 & 45.93 & 55.64 & 35.45 & 56.87 \\
\rowcolor{oursrow} \ours (SFT init.) & FD & 31.97 & 75.32 & 80.47 & 57.38 & 29.26 & \bd{82.19} & 84.26 & \bd{66.99} & 65.58 & 46.24 & 54.70 & 31.06 & 67.49 \\
\rowcolor{oursrow} \ours (DPO init.) & FD & \bd{31.00} & \bd{76.52} & \bd{83.57} & \bd{60.15} & \bd{26.25} & 79.65 & \bd{84.74} & 64.88 & \bd{58.86} & \bd{53.55} & \bd{64.51} & \bd{36.25} & \bd{69.85} \\
\bottomrule
\end{tabular}
\end{table*}

\noindent\textbf{Benchmark and Metrics.}
We evaluate \ours on TravelUAV~\cite{wang2025travelUAV}, a benchmark for aerial VLN built on AirSim and Unreal Engine~\cite{shah2018airsim}.
Its training set contains 7,611 routes in 11 scenes, and its test set contains 1,418 routes in seen environments (Seen), 629 routes with unseen objects (Unseen Object), and 958 routes in unseen environments (Unseen Map).
\ours collects experience only on the training routes.
An episode succeeds if the UAV lands within 20\,m of the target within 200 actions.
Following~\cite{xu2026aerodpo}, we report the navigation error (NE, in meters), success rate (SR), oracle success rate (OSR), and success weighted by path length (SPL), together with the collision rate (CR).
All design choices were made on training routes held out in preliminary experiments, and every round of \ours is evaluated on all test routes.
Since closed-loop flight is stochastic, evaluating the same policy twice changes the outcome of about a quarter of the routes and its overall SR by up to about two points, so we compare policies route by route and treat smaller differences as noise.
The simulator pauses at observation boundaries during data collection so that states can be saved and restored exactly, whereas all evaluations follow the native protocol without pauses.

\noindent\textbf{Baselines.}
We compare with two groups of methods.
TravelUAV~\cite{wang2025travelUAV}, NavFoM~\cite{zhang2026navfom}, LongFly~\cite{jiang2025longfly}, and NeuroKalman~\cite{tang2026neurokalman} use the L1 assistant guidance of TravelUAV, whereas AerialVLA~\cite{xu2026aerialvla}, RefineFly~\cite{wang2026refinefly}, and AeroDPO~\cite{xu2026aerodpo} use fuzzy direction guidance, as \ours does.
AeroDPO is the closest baseline.
It first trains a 2B-parameter policy on the expert demonstrations of AerialVLA by supervised fine-tuning, which we denote SFT, and then post-trains it by preference optimization on corrections derived from collision rollbacks.
For a controlled comparison, \ours (SFT init.) starts from the same SFT policy, and we evaluate the released SFT and AeroDPO checkpoints under the same protocol as \ours, comparing them route by route.

\noindent\textbf{Implementation Details.}
Table~\ref{tab:setup} lists the compared policies.
To test whether \ours also improves a policy that has already been post-trained, we further start it from AeroDPO, denoted \ours (DPO init.).
In both cases, $\pi_0$ is the initial policy, and the states that need no repair contribute only their observations, labeled with the outputs of the current policy.
The supplementary material gives all settings and the cost of each round.

\subsection{Main Results}
\label{sec:exp_main}

\noindent\textbf{Comparison with Existing Methods.}
Table~\ref{tab:main} compares \ours after three rounds with existing methods.
\ours (SFT init.) and AeroDPO post-train the same SFT policy without additional human labels, so their difference reflects the post-training alone.
\ours raises the overall SR 10.6 points above AeroDPO, and the gain holds route by route, as \ours succeeds on 649 test routes on which AeroDPO fails while failing on only 330 on which AeroDPO succeeds.
The gain concentrates where the loop collects experience.
On Seen and Unseen Object, \ours outperforms all prior methods on every metric, raising the SR of AeroDPO by 15.3 and 15.7 points and reducing its NE by nearly 40\%, whereas on Unseen Map, whose environments the loop never visits, it raises the SR of SFT by 9.4 points but only matches AeroDPO in SR, with a lower SPL.
SPL rises less than SR, as many repairs climb over or detour around obstacles and lengthen the path.
\ours (DPO init.) improves every metric on all three splits and raises the overall SR of AeroDPO by another 13.0 points to 69.9\%, so repairs verified by the policy remain effective after preference optimization and complement it.
On Unseen Map, the loop adds a similar gain to either initial policy, 9.4 and 7.6 points, so the final level in unseen environments is set mainly by the initial policy.

\begin{table}[t]
\centering
\footnotesize
\setlength{\tabcolsep}{3.1pt}
\caption{\textbf{Results across rounds.} Seen, UO, and UM give the SR on each split; SR, SPL, and CR are computed over all test routes. Train SR: SR on the training routes. Round 0 is the initial policy. The best result of each setting is in bold.}
\label{tab:rounds}
\begin{tabular}{lccccccc}
\toprule
Round & Seen & UO & UM & SR\up & SPL\up & CR\down & Train SR\up \\
\midrule
\multicolumn{8}{l}{\textit{\ours (SFT init.)}} \\
0 & 48.38 & 56.44 & 36.85 & 46.39 & 37.37 & 38.64 & 45.24 \\
1 & 62.62 & 67.89 & 43.42 & 57.60 & 46.96 & 29.48 & 61.41 \\
2 & 73.91 & 79.01 & 45.82 & 66.02 & \bd{51.96} & 21.26 & \bd{71.41} \\
\rowcolor{oursrow} 3 & \bd{75.32} & \bd{82.19} & \bd{46.24} & \bd{67.49} & 51.00 & \bd{19.10} & 70.85 \\
\midrule
\multicolumn{8}{l}{\textit{\ours (DPO init.)}} \\
0 & 60.01 & 66.45 & 45.93 & 56.87 & 45.20 & 28.95 & 56.21 \\
1 & 69.96 & 73.93 & 46.76 & 63.39 & 49.84 & 22.46 & 66.03 \\
2 & 72.50 & 78.54 & 48.85 & 66.22 & 50.07 & 19.07 & 70.84 \\
3 & \bd{76.52} & \bd{79.65} & \bd{53.55} & \bd{69.85} & \bd{53.52} & \bd{17.90} & \bd{72.78} \\
\bottomrule
\end{tabular}
\end{table}

\begin{figure}[t]
\centering
\includegraphics[width=\linewidth]{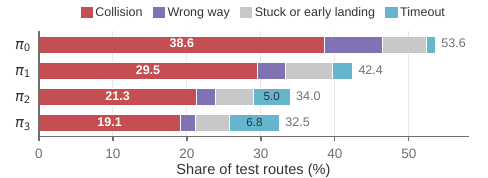}
\caption{\textbf{Failure types across rounds.} Share of test routes that end in each type of failure for the policy of every round of \ours (SFT init.), with the total failure rate after each bar.}
\label{fig:failures}
\end{figure}

\noindent\textbf{Improvement across Rounds.}
Table~\ref{tab:rounds} reports every round.
For \ours (SFT init.), the test SR rises from 46.4\% to 57.6\%, 66.0\%, and 67.5\%, and the SR on the training routes follows the same trend, so the repairs generalize beyond the anchored states.
The gain per round decreases from 11.2 to 8.4 and 1.5 points, the last within the noise, as each round removes the failures that its verifier can repair.
The failures that remain also change in kind (Fig.~\ref{fig:failures}).
Collisions and flights away from the target decrease in every round, whereas timeouts grow from 1.2\% to 6.8\% of the routes, so the bottleneck shifts from avoiding obstacles to reaching the target in time.
\ours (DPO init.) follows the same trend, and the gap between the two initializations shrinks from 10.5 to 2.4 points, almost all of it on Unseen Map, so in the environments that the loop experiences, the final result depends little on which policy it starts from.

\subsection{Ablation Study}
\label{sec:exp_ablation}

\begin{table*}[t]
\centering
\footnotesize
\setlength{\tabcolsep}{3.2pt}
\caption{\textbf{Ablation study on all test routes.} Variants in (a) and (c) are compared with the first round and those in (b) with the second. In (a), unverified variants label every anchor without flying the label, verified variants keep only the labels that the policy completes, and the two-action variant scores the candidates after two actions instead. Paired significance tests are in the supplementary material.}
\label{tab:ablation}
\begin{subtable}[t]{0.36\linewidth}
\centering
\caption{Verification in the first round}
\label{tab:abl_verify}
\begin{tabular}{lcccc}
\toprule
Source & Verified & SR\up & SPL\up & CR\down \\
\midrule
Potential field & -- & 39.75 & 31.92 & 25.01 \\
Rule-based proposals & -- & 49.23 & 37.58 & 24.08 \\
Rule-based proposals & \checkmark & 61.59 & 48.88 & 24.92 \\
Sampled actions & -- & 50.62 & 41.01 & 34.93 \\
Sampled actions & \checkmark & 51.65 & 42.95 & 36.33 \\
Both & 2 actions & 55.76 & 47.00 & 33.72 \\
\rowcolor{oursrow} Both (ours) & \checkmark & 57.60 & 46.96 & 29.48 \\
\bottomrule
\end{tabular}
\end{subtable}\hfill
\begin{subtable}[t]{0.31\linewidth}
\centering
\caption{Recursion}
\label{tab:abl_recursion}
\begin{tabular}{lccc}
\toprule
Variant & SR\up & SPL\up & CR\down \\
\midrule
Round 1 & 57.60 & 46.96 & 29.48 \\
Stale labels & 60.30 & 48.81 & 28.22 \\
$2{\times}$ anchors, once & 62.98 & 51.99 & 26.10 \\
w/o replay & 65.68 & 52.93 & 22.60 \\
\rowcolor{oursrow} Round 2 (ours) & 66.02 & 51.96 & 21.26 \\
\bottomrule
\end{tabular}
\end{subtable}\hfill
\begin{subtable}[t]{0.29\linewidth}
\centering
\caption{Repair learning}
\label{tab:abl_modules}
\begin{tabular}{lccc}
\toprule
Variant & SR\up & SPL\up & CR\down \\
\midrule
w/o retention & 57.46 & 47.28 & 32.01 \\
\quad + DPO loss & 58.04 & 47.26 & 30.67 \\
States from $\pi_0$ & 58.97 & 48.39 & 30.00 \\
Advantage weights & 53.93 & 43.98 & 33.87 \\
\rowcolor{oursrow} Round 1 (ours) & 57.60 & 46.96 & 29.48 \\
\bottomrule
\end{tabular}
\end{subtable}
\end{table*}

\begin{figure*}[t]
\centering
\includegraphics[width=\linewidth]{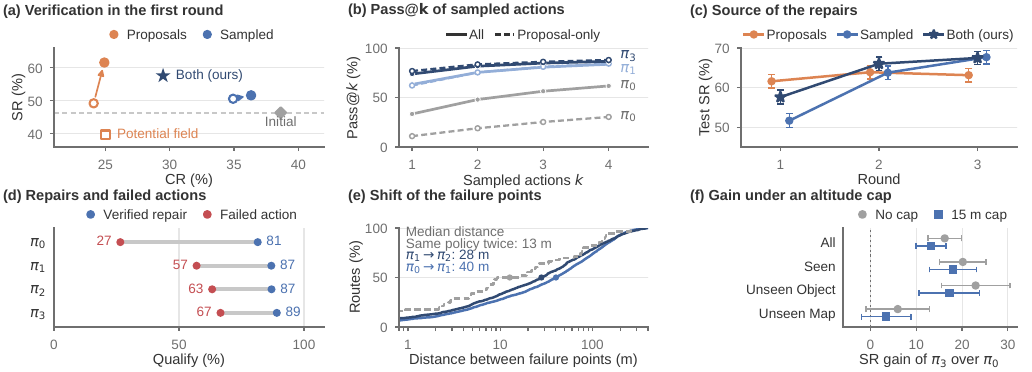}
\caption{\textbf{Analysis of the design of \ours.} (a)~SR and CR after the first round when the labels are verified by the policy (filled) or not (hollow), as in Table~\ref{tab:ablation}(a). (b)~Pass@$k$ of newly sampled actions of each policy at the reproducible anchors of the first round, at all of them (solid) and at those that only a rule-based proposal repaired (dashed). (c)~Test SR when a round learns from the rule-based proposals, the sampled actions, or both, each single-source round starting from the previous round with both sources. (d)~At the anchors accepted in the first round, the verified repair and the action at which $\pi_0$ failed, each continued by a later policy ($\pi_0$ continues the repair again). (e)~Distance between the failure points of two flights on the training routes on which both fail. (f)~SR gain of $\pi_3$ over $\pi_0$ on 999 test routes with and without an altitude cap. Error bars are 95\% confidence intervals, paired in (f).}
\label{fig:ablation}
\end{figure*}

\begin{figure*}[t]
\centering
\includegraphics[width=\linewidth]{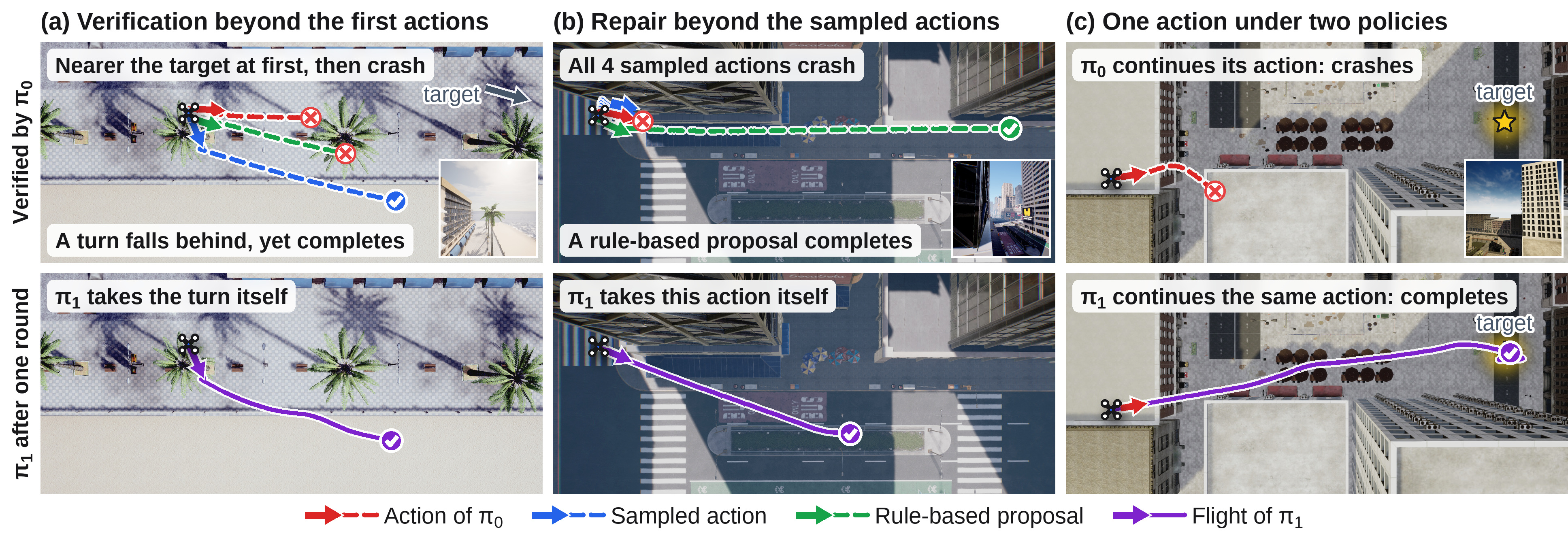}
\caption{\textbf{Case study of \ours.} From a state before $\pi_0$ crashes, $\pi_0$ flies on from each candidate action (top), and $\pi_1$ flies after one round (bottom). (a)~Flying toward the target looks best at first but crashes, whereas a turn that falls behind completes the flight, so a repair can be judged only by flying on. (b)~The sampled actions nearly repeat the crashing action, so only a rule-based proposal repairs it, and $\pi_1$ learns this repair. (c)~The same action crashes when $\pi_0$ flies on but not when $\pi_1$ does, so every round must judge its repairs anew.}
\label{fig:qualitative}
\end{figure*}

Table~\ref{tab:ablation} and Fig.~\ref{fig:ablation} examine the insights behind \ours first and the designs of its modules afterwards.

\noindent\textbf{Verification by the Policy.}
Verification by the policy is what turns candidate actions into useful supervision (Table~\ref{tab:ablation}(a)).
Labeling every anchor with the potential-field action, as DAgger~\cite{ross2011dagger} would with a rule-based expert, lowers the SR of the initial policy by 6.6 points, and a random rule-based proposal raises it by only 2.8 points, as the policy trades collisions for getting stuck or flying away from the target (Fig.~\ref{fig:ablation}(a)).
Once the policy must complete the same proposals, the SR rises by 12.4 points to 61.6\%.
For its own sampled actions, which stay close to its behavior, verification adds only 1.0 point, so the benefit of verification grows with how far a source departs from the policy.
This benefit comes from flying the candidates on rather than from the checks of Sec.~\ref{sec:method_cvr}, without which the SR is a similar 58.4\%.
Verification must also follow the flight beyond the first actions, as scoring the candidates after two actions instead leaves 4.2 points more collisions, and the candidate nearest to the target after two actions later fails or stalls at nearly a quarter of the anchors with a repair (Fig.~\ref{fig:qualitative}(a)).
Repairs mostly hold beyond the horizon, as $\pi_0$, flying on from those of the first round until it lands, collides, or runs out of actions, reaches the target in 57\% of episodes it had failed.

\noindent\textbf{Sources of Candidates.}
The two sources of candidates complement each other, and their roles shift as the policy learns (Fig.~\ref{fig:ablation}(b, c)).
The failing policy often holds a repair that it does not choose, as sampled actions supply 44\% of the repairs accepted in the first round, and 78\% of them still qualify when $\pi_0$ flies them again, against 38\% for new samples that nearly repeat the failed action, so they are not lucky continuations.
At the 21\% of the anchors whose failure reproduces that only a rule-based proposal repaired, however, a sampled action of $\pi_0$ qualifies at only 11\%, whereas after one round of learning from both sources, a sampled action of $\pi_1$ qualifies at 62\%, as often as at the other anchors, and at 58\% of the 42 such anchors in route groups held out from training, against 14\% for $\pi_0$.
The repairs learned from the proposals thus enter the policy's sampled actions, and the leading source changes across rounds (Fig.~\ref{fig:ablation}(c)).
The proposals alone lead in the first round, as nearly all of their repairs change a single component of the action and thus teach a consistent correction, and both sources together lead in the second.
No single source suffices in every round, so \cvr verifies both and keeps the best candidate.

\noindent\textbf{Necessity of Recursion.}
A new round is needed because the failures move, not because the repairs expire (Table~\ref{tab:ablation}(b)).
Retraining $\pi_1$ on the repairs of the first round, without anchoring or verifying anew, reaches 60.3\%, and collecting twice as many anchors from $\pi_0$ and training once, with as many updates as two rounds, reaches 63.0\%, both below the 66.0\% of the second round and with more collisions, whereas the second round without replay still reaches 65.7\%.
Yet the repairs remain valid, as those verified in the first round still qualify at 87\% to 89\% of their anchors when later policies continue them, compared with 81\% when $\pi_0$ continues them again (Fig.~\ref{fig:ablation}(d)).
What changes is where the policy fails.
The action at which $\pi_0$ failed qualifies at 57\% to 67\% of these anchors under later policies, against 27\% under $\pi_0$, and on the training routes that two consecutive policies both fail, their failure points lie a median of 40\,m apart after the first round and 28\,m after the second, compared with 13\,m when the same policy flies twice (Fig.~\ref{fig:ablation}(e)).

\noindent\textbf{Retentive Repair Learning.}
Behavior retention keeps the repairs from changing normal flight (Table~\ref{tab:ablation}(c)).
Training on the repairs alone raises the CR from 29.5\% to 32.0\%, as the policy turns in 19.6\% instead of 15.8\% of its actions, drifting toward the 30\% of the repairs.
Taking the retained states from flights of $\pi_0$ instead of AeroDPO gives 59.0\%, so the retention does not rely on another method.
Weighting every qualifying candidate by its group-normalized advantage, as in GRPO~\cite{shao2024deepseekmath}, is 3.7 points below imitating the best candidate alone, suggesting that one consistent repair per state teaches more than several competing ones.
Adding the DPO loss against the original action gives a similar 58.0\%, while the DPO loss alone collapses the policy, so the gain comes from the verified repairs rather than the objective.

\noindent\textbf{Flight Altitude.}
Many verified repairs climb over obstacles, and the median peak altitude of a test flight rises from 10\,m for $\pi_0$ to 36\,m for $\pi_3$, yet with the altitude capped at 15\,m above the start on 999 test routes, $\pi_3$ keeps about four fifths of its gain over $\pi_0$, 13.2 of 16.2 points (Fig.~\ref{fig:ablation}(f)).

\subsection{Case Study}
\label{sec:exp_case}

\noindent\textbf{Missed Repairs.}
Fig.~\ref{fig:qualitative} shows three anchors of the first round, with the front camera in the insets.
Panels (a) and (b) reveal two ways in which a policy misses a repair that exists.
In (a), the policy samples the repair but does not choose it.
After two actions, a sharp turn is about 6\,m behind the actions flying toward the target, yet those actions crash into a palm tree and only the turn completes the flight.
A repair can thus cost progress before it pays off, so a criterion on the first actions would discard exactly the behavior that the policy lacks.
In (b), the policy does not even sample the repair, as its sampled actions nearly repeat the action that crashes.
Only one of the three rule-based proposals completes the flight, so rules can supply a repair, but only the flight of the policy can identify it.
In both cases, $\pi_1$ takes the repair by itself after one round, as do most of its new sampled actions, so each round shifts what the next one can sample.

\noindent\textbf{One Action under Two Policies.}
Fig.~\ref{fig:qualitative}(c) shows that the value of an action is not fixed.
The action at which $\pi_0$ crashes completes the flight when $\pi_1$ flies on from it, as it does at most anchors accepted in the first round (Fig.~\ref{fig:ablation}(d)).
Whether an action repairs a failure thus depends on the policy that continues the flight, so a label judged under an earlier policy or an expert may not hold for the current one.
This is why \ours lets the current policy judge its own repairs and verifies them anew in every round.

\section{Conclusion}
\label{sec:conclusion}

In this paper, we have presented \ours, a recursive self-improvement framework for aerial VLN in which a failing policy judges its own repairs by flying on from them.
Verified in this way, the sampled actions of the policy and simple rule-based proposals become reliable supervision, and since repairs remain valid while failures move, every round anchors and verifies anew.
On TravelUAV, three rounds raise the SR of a 2B-parameter policy from 46.4\% to 67.5\% and reduce collisions by half without additional human labels or changes at deployment.

{
    \small
    \bibliographystyle{ieeenat_fullname_unsrt}
    \bibliography{references}
}

\floatsep=\aerorsiFloatsep \textfloatsep=\aerorsiTextFloatsep \intextsep=\aerorsiIntextsep
\dblfloatsep=\aerorsiDblFloatsep \dbltextfloatsep=\aerorsiDblTextFloatsep
\setcounter{table}{0}
\setcounter{algorithm}{0}
\maketitlesupplementary
\renewcommand{\thetable}{S\arabic{table}}
\renewcommand{\thealgorithm}{S\arabic{algorithm}}
\appendix

This supplementary material is organized as follows.
\begin{itemize}[leftmargin=*,itemsep=0pt,topsep=2pt]
\item Sec.~\ref{sec:s_impl} gives the implementation details of \ours: its settings, the rule-based proposals, the verification of an anchor (Algorithm~\ref{alg:s_cvr}), the ablation variants, and the evaluation and cost.
\item Sec.~\ref{sec:s_results} gives the failure types of both initializations and the paired significance tests behind the comparisons of the main paper.
\item Sec.~\ref{sec:s_analysis} gives further analyses behind Fig.~4 of the main paper, in the order of its ablation study.
\end{itemize}

\section{Implementation Details}
\label{sec:s_impl}

\subsection{Settings}
Table~\ref{tab:s_settings} lists the settings of \ours, which are shared by both initializations and all three rounds.
A fifth of the route groups of the training split is held out from training: the policies still fly these routes and \afa anchors their failures, but their repairs only serve to test whether repairs transfer to route groups that the policy has not learned from (Secs.~\ref{sec:s_sources} and~\ref{sec:s_move}).
The retention states of \rrl come from successful flights of AeroDPO on the training routes and are labeled with the greedy outputs of the current policy; taking them from the flights of $\pi_0$ instead gives a similar SR (Table~4(c) of the main paper).

\begin{table}[h]
\centering
\footnotesize
\setlength{\tabcolsep}{3.5pt}
\caption{\textbf{Settings of \ours.} They are shared by both initializations and all rounds.}
\label{tab:s_settings}
\begin{tabular}{@{}l>{\raggedright\arraybackslash}p{0.80\linewidth}@{}}
\toprule
Component & Setting \\
\midrule
Policy & Qwen3-VL-2B; $384{\times}768$ mosaic of the front and downward views; forward, vertical, and yaw components with 99 bins each, plus landing \\
Training & LoRA of rank 64 on all linear layers of the language model, and the visual merger; 2 epochs per round; learning rate $2{\times}10^{-5}$; batch size 16 \\
\afa & Offsets of 1, 2, 4, 8, and 16 actions; 2,400 anchors per round; horizon of $\max(8, \delta{+}4)$ actions for offset $\delta$, at most 20 \\
\cvr & $M{=}4$ samples at temperature 1.0 and three rule-based proposals; $\lambda_f{=}100$, $\lambda_s{=}20$; minimum progress of 4\,m; margin $m{=}4$\,m; two confirmations at most \\
\rrl & $\lambda_n{=}0.2$ on 1,990 retention states; replay share $\rho{=}30\%$ from round 2 \\
\bottomrule
\end{tabular}
\end{table}

\subsection{Rule-based Proposals}
\cvr adds three rule-based proposals to the sampled actions, each computed from the anchor state alone.
The \emph{potential field} attracts the UAV to the target and repels it from the obstacles in the depth map observed at the anchor.
Its \emph{translation} variant keeps the forward command of the original action, takes the vertical command of the potential field, and limits the yaw to 0.224\,rad, below the 0.25\,rad at which an action turns in place instead of translating.
The \emph{climb} keeps the forward and yaw commands of the original action and rises 2\,m more, by at most 5\,m in one action.
The proposals are deliberately simple and become useful only after verification (Table~4(a) of the main paper).

\subsection{Verification}
Algorithm~\ref{alg:s_cvr} details how \cvr verifies one anchor.
A flight fails if it collides, lands prematurely, or produces an invalid output, and a candidate qualifies if its flight reaches the end of the horizon without failure and with at least 4\,m of progress, or if it lands successfully.
The input check discards an anchor at which the restored observation no longer leads the policy to its original action.
The reproduction check keeps an anchor only if the original action is indeed worse than a candidate, and the confirmation check accepts a candidate only if it qualifies in two independent flights.

\begin{algorithm}[t]
\caption{Verification of one anchor by \cvr}
\label{alg:s_cvr}
\footnotesize
\begin{algorithmic}[1]
\Require policy $\pi_k$; anchor state $s$ with observation $o$, original action $a^{-}$, and horizon $H$
\Ensure a verified repair $(o, c^{\star})$, or none
\If{the greedy action of $\pi_k$ at $o$ is not $a^{-}$} \Comment{input check}
\State \Return none
\EndIf
\State $C \gets$ up to $M$ distinct samples of $\pi_k(\cdot \mid o)$ and the three rule-based proposals, other than $a^{-}$
\For{each $c \in \{a^{-}\} \cup C$}
\State Restore $s$, execute $c$, and let $\pi_k$ fly on greedily for up to $H$ actions
\State Score the flight by $S_k(s, c)$ (Eq.~(3) of the main paper)
\EndFor
\State $Q \gets$ the qualifying candidates in $C$, by decreasing $S_k$
\If{$a^{-}$ does not fail \textbf{and} ($Q = \emptyset$ \textbf{or} $S_k(s, Q_1) - S_k(s, a^{-}) < m$)}
\State \Return none \Comment{reproduction check}
\EndIf
\For{each $c$ among the first two of $Q$} \Comment{confirmation check}
\State Restore $s$ and fly $c$ again in the same way
\If{this flight qualifies}
\State \Return $(o, c)$
\EndIf
\EndFor
\State \Return none
\end{algorithmic}
\end{algorithm}

\subsection{Variants of the Ablation Study}
Three variants in Table~4 of the main paper are set up as follows.
Without the two checks, \cvr accepts the best qualifying candidate of every anchor after one flight, giving 1,374 repairs instead of 837 with the same 354 updates.
Without replay, round 2 trains $\pi_1$ on its new repairs and the retention states with the same 444 updates.
The DPO variant pairs the 837 repairs with the original action as the rejected one, with $\pi_0$ as the reference, $\beta{=}0.1$, and the learning rate and 354 updates of round 1, without retention states; with the DPO loss alone, the outputs degenerate and the SR falls to 0.1\%, whereas adding the likelihood of the repair with weight 1 gives 58.04\%.

\subsection{Evaluation and Cost}
All policies are evaluated on the 3,005 test routes of TravelUAV with the native actions and termination of the benchmark, and an episode interrupted by a simulator error is flown again from its start.
A round of \ours costs about 100 GPU hours (Table~\ref{tab:s_cost}): flying the 7,611 training routes takes 17 to 20 hours and verifying the anchors about 6 hours on four GPUs, and training takes 11 to 23 minutes on one GPU.

\begin{table}[t]
\centering
\footnotesize
\setlength{\tabcolsep}{4pt}
\caption{\textbf{Simulator cost of each round} of \ours (SFT init.). Round $k$ flies the training routes with $\pi_{k-1}$.}
\label{tab:s_cost}
\begin{tabular}{lccc}
\toprule
Round & 1 & 2 & 3 \\
\midrule
\multicolumn{4}{l}{\textit{Flights of the training routes}} \\
Flights & 7,611 & 7,611 & 7,611 \\
Actions (thousands) & 345 & 376 & 435 \\
\midrule
\multicolumn{4}{l}{\textit{Verification}} \\
Flights of candidates & 15,237 & 13,975 & 12,451 \\
Confirmation flights & 1,484 & 1,342 & 1,165 \\
Actions (thousands) & 141 & 125 & 116 \\
\bottomrule
\end{tabular}
\end{table}

\section{Additional Results}
\label{sec:s_results}

\subsection{Failure Types across Rounds}
Table~\ref{tab:s_outcomes} extends Fig.~3 of the main paper to both initializations.
From either initial policy, collisions and flights away from the target decrease over the rounds, so the gain comes mainly from flights that no longer crash or lose the target.

\begin{table}[t]
\centering
\footnotesize
\setlength{\tabcolsep}{4.5pt}
\caption{\textbf{Failure types across rounds.} Share (\%) of all test routes that end in each type of failure; the other routes succeed (Table~3 of the main paper).}
\label{tab:s_outcomes}
\begin{tabular}{lccccc}
\toprule
Round & Collision & Wrong way & Stuck & Early landing & Timeout \\
\midrule
\multicolumn{6}{l}{\textit{\ours (SFT init.)}} \\
0 & 38.6 & 7.8 & 4.2 & 1.8 & 1.2 \\
1 & 29.5 & 3.8 & 4.2 & 2.2 & 2.8 \\
2 & 21.3 & 2.6 & 2.8 & 2.3 & 5.0 \\
3 & 19.1 & 2.1 & 2.7 & 1.9 & 6.8 \\
\midrule
\multicolumn{6}{l}{\textit{\ours (DPO init.)}} \\
0 & 29.0 & 5.5 & 4.2 & 1.3 & 3.2 \\
1 & 22.5 & 3.5 & 3.2 & 2.6 & 4.9 \\
2 & 19.1 & 3.4 & 3.6 & 2.6 & 5.1 \\
3 & 17.9 & 3.1 & 2.9 & 2.9 & 3.3 \\
\bottomrule
\end{tabular}
\end{table}

\subsection{Paired Significance Tests}
Table~\ref{tab:s_paired} tests the comparisons of the main paper route by route.
For two runs on the same test routes, with $b$ routes on which only the first succeeds and $c$ routes on which only the second does, we report the difference in SR and $z=(b-c)/\sqrt{b+c}$ of the McNemar test, where $|z|>1.96$ means $p<0.05$.
For the two comparisons that the main paper makes by collisions, the CR rises by 4.23 points with the two-action score ($z{=}4.2$) and by 2.53 points without retention ($z{=}2.6$).

\begin{table}[t]
\centering
\footnotesize
\setlength{\tabcolsep}{4pt}
\caption{\textbf{Paired tests on the test routes.} Difference in SR (points) and $z$ over the routes common to both runs, for \ours (SFT init.) unless stated.}
\label{tab:s_paired}
\begin{tabular}{lcc}
\toprule
Comparison & $\Delta$SR & $z$ \\
\midrule
\multicolumn{3}{l}{\textit{Rounds and baselines}} \\
Round 1 vs.\ round 0 & +11.21 & 10.4 \\
Round 2 vs.\ round 1 & +8.42 & 8.1 \\
Round 3 vs.\ round 2 & +1.46 & 1.6 \\
Round 3 vs.\ round 0 & +21.10 & 18.5 \\
Round 3 vs.\ AeroDPO & +10.62 & 10.2 \\
DPO init., round 3 vs.\ AeroDPO & +12.98 & 12.7 \\
DPO init.\ vs.\ SFT init., round 3 & +2.36 & 2.5 \\
\midrule
\multicolumn{3}{l}{\textit{Ablations}} \\
Potential field vs.\ round 0 & $-$6.64 & $-$6.2 \\
Rule-based proposals, unverified vs.\ verified & $-$12.37 & $-$11.5 \\
Without the two checks vs.\ round 1 & +0.77 & 0.8 \\
Rule-based proposals vs.\ both, round 2 & $-$2.13 & $-$2.2 \\
Sampled actions vs.\ both, round 2 & $-$2.23 & $-$2.3 \\
Stale labels vs.\ round 2 & $-$5.73 & $-$5.5 \\
$2{\times}$ anchors, once vs.\ round 2 & $-$3.06 & $-$3.2 \\
w/o replay vs.\ round 2 & $-$0.33 & $-$0.4 \\
Advantage weights vs.\ round 1 & $-$3.70 & $-$3.7 \\
+ DPO loss vs.\ w/o retention & +0.57 & 0.6 \\
\bottomrule
\end{tabular}
\end{table}

\section{Additional Analysis}
\label{sec:s_analysis}

\subsection{Statistics of the Loop}
\label{sec:s_loop}
Table~\ref{tab:s_loop} summarizes the collection of each round.
As the policy improves, fewer anchors yield a repair, whereas the best of four sampled actions still qualifies at about two thirds of the anchors whose failure reproduces, and a growing share of the repairs comes from the sampled actions.
From the second round on, about half of these sampled repairs change a single component of the failed action, against 31\% in the first, closer to the rule-based proposals, nearly all of whose repairs do.
Repairs are found both just before and long before a failure: in the first round, a repair is accepted at 64\% of the anchors one action before a collision and at 56\% of those 16 actions before.

\begin{table}[t]
\centering
\footnotesize
\setlength{\tabcolsep}{4pt}
\caption{\textbf{Statistics of the loop} of \ours (SFT init.) unless stated. Anchors: those evaluated of the 2,400 drawn.}
\label{tab:s_loop}
\begin{tabular}{lccc}
\toprule
Round & 1 & 2 & 3 \\
\midrule
Anchors & 2,390 & 2,390 & 2,399 \\
\midrule
\multicolumn{4}{l}{\textit{Share of all anchors (\%)}} \\
Repair accepted & 43.8 & 38.3 & 34.3 \\
Repair accepted, DPO init. & 46.1 & 38.8 & 33.9 \\
\midrule
\multicolumn{4}{l}{\textit{Share of the anchors whose failure reproduces (\%)}} \\
Best of four samples qualifies & 66.2 & 68.3 & 68.3 \\
Only a rule-based proposal repairs & 21.0 & 16.4 & 16.5 \\
\midrule
\multicolumn{4}{l}{\textit{Share of the accepted repairs (\%)}} \\
From sampled actions & 43.9 & 51.9 & 54.4 \\
\quad of which change one component & 30.7 & 50.5 & 47.9 \\
\bottomrule
\end{tabular}
\end{table}

\subsection{Sources of Candidates}
\label{sec:s_sources}
Table~\ref{tab:s_candidates} gives how often the candidates of each type qualify at the anchors whose failure reproduces.
The rule-based proposals qualify at about half of these anchors, so they are useful candidates but unreliable labels without verification, and the sampled actions qualify more often as the policy improves.
The repairs taken from sampled actions are not lucky continuations: at the anchors accepted in the first round, 77.6\% of them qualify again when $\pi_0$ flies them once more (Table~\ref{tab:s_repairs}), whereas new samples of $\pi_0$ that nearly repeat the failed action, within two bins in every component, qualify at 37.8\%.

\begin{table}[t]
\centering
\footnotesize
\setlength{\tabcolsep}{4pt}
\caption{\textbf{Candidates that qualify} (\%) at the anchors whose failure reproduces. Identical samples of an anchor count once.}
\label{tab:s_candidates}
\begin{tabular}{lcccc}
\toprule
Round & Sampled actions & Potential field & Translation & Climb \\
\midrule
1 & 39.2 & 60.7 & 59.3 & 47.6 \\
2 & 44.3 & 55.4 & 59.6 & 46.2 \\
3 & 51.0 & 63.6 & 59.0 & 51.2 \\
\bottomrule
\end{tabular}
\end{table}

Table~\ref{tab:s_groups} extends Fig.~4(b) of the main paper to all groups of the 1,450 anchors of the first round whose failure reproduces.
The groups are defined by the samples drawn during collection and the rates by four new samples of each policy, so the two do not share the same draws.
The anchors that only a rule-based proposal repaired gain the most, also in the route groups held out from training, and at these anchors the qualifying samples of later policies come to resemble the proposals: the share that climbs by at least 10 bins grows from 12\% for $\pi_0$ to 42\% for $\pi_1$ and 57\% for $\pi_3$.

\begin{table}[t]
\centering
\footnotesize
\setlength{\tabcolsep}{3.4pt}
\caption{\textbf{Pass@$k$ of new samples} (\%) at the 1,450 reproducible anchors of the first round, grouped by what repaired them during collection.}
\label{tab:s_groups}
\begin{tabular}{lcccccc}
\toprule
& \multicolumn{2}{c}{$\pi_0$} & \multicolumn{2}{c}{$\pi_1$} & \multicolumn{2}{c}{$\pi_3$} \\
\cmidrule(lr){2-3}\cmidrule(lr){4-5}\cmidrule(lr){6-7}
Anchors repaired by & @1 & @4 & @1 & @4 & @1 & @4 \\
\midrule
All (100\%) & 33.4 & 61.7 & 63.2 & 83.6 & 73.8 & 86.6 \\
A sampled action (66\%) & 45.6 & 79.9 & 71.7 & 90.9 & 80.4 & 92.2 \\
Only a proposal (21\%) & 11.0 & 30.4 & 62.1 & 84.3 & 76.7 & 87.9 \\
Neither (13\%) & 6.5 & 18.9 & 20.9 & 44.3 & 34.6 & 55.1 \\
\midrule
\multicolumn{7}{l}{\textit{Held-out route groups}} \\
All & 41.8 & 74.9 & 68.7 & 91.8 & 79.0 & 92.7 \\
Only a proposal (42) & 13.7 & 33.3 & 57.7 & 85.7 & 76.2 & 85.7 \\
\bottomrule
\end{tabular}
\end{table}

\subsection{Repairs and Failures across Rounds}
\label{sec:s_move}
Table~\ref{tab:s_repairs} flies the 1,048 repairs accepted in the first round again from their anchors, each continued by a later policy, as in Fig.~4(d) of the main paper.
The repairs remain valid under every later policy, from either source and in the route groups held out from training, whereas the action at which $\pi_0$ failed passes increasingly often.
What changes instead is where the policy fails (Table~\ref{tab:s_points}): on the training routes that two consecutive policies both fail, the two failure points lie much farther apart than when one policy flies the same routes twice.
Most repairs also hold to the end of the episode.
From each of the 1,048 anchors accepted in the first round, $\pi_0$ executes the verified repair and flies on until it lands, collides, or reaches the limit of 200 actions, with the simulator paused at observation boundaries as in collection.
It reaches the target in 57.2\% of these episodes, 58.2\% for repairs from sampled actions and 56.5\% for those from rule-based proposals, and in 53.6\% at the 604 anchors whose failure reproduces.
The flights that fail again do so a median of 14 actions after the original failure point.

\begin{table}[t]
\centering
\footnotesize
\setlength{\tabcolsep}{4pt}
\caption{\textbf{Verified repairs under later policies.} Share (\%) of the 1,048 anchors accepted in the first round at which the repair, or the action at which $\pi_0$ failed, qualifies.}
\label{tab:s_repairs}
\begin{tabular}{lcccc}
\toprule
Continued by & $\pi_0$ again & $\pi_1$ & $\pi_2$ & $\pi_3$ \\
\midrule
Verified repair & 81.5 & 86.9 & 87.0 & 89.1 \\
\quad held out (211) & 82.5 & 88.2 & 91.9 & 92.9 \\
\quad from sampled actions (460) & 77.6 & 83.3 & 85.4 & 88.0 \\
\quad from proposals (588) & 84.5 & 89.8 & 88.2 & 90.0 \\
Failed action & 26.5 & 57.0 & 63.3 & 66.6 \\
\bottomrule
\end{tabular}
\end{table}

\begin{table}[t]
\centering
\footnotesize
\setlength{\tabcolsep}{4.5pt}
\caption{\textbf{Distance between the failure points} of two flights on the routes that both fail: share (\%) within 10\,m and over 50\,m, and the median. One policy flown twice is an earlier policy on 384 held-out routes.}
\label{tab:s_points}
\begin{tabular}{lcccc}
\toprule
Flights & Routes & Within 10\,m & Over 50\,m & Median \\
\midrule
$\pi_0$, then $\pi_1$ & 2,125 & 27.9 & 44.8 & 40.2\,m \\
$\pi_1$, then $\pi_2$ & 1,333 & 33.0 & 37.6 & 28.0\,m \\
One policy twice & 56 & 50.0 & 30.4 & 12.7\,m \\
\bottomrule
\end{tabular}
\end{table}

\subsection{Flight Altitude}
\label{sec:s_altitude}
The median peak altitude of a test flight of \ours (SFT init.) rises from 10.0\,m in round 0 to 15.6, 29.1, and 36.2\,m in rounds 1 to 3.
To test how much of the gain this explains, we evaluate $\pi_0$, $\pi_1$, and $\pi_3$ on 999 test routes drawn from the three splits with the altitude capped at 15\,m above the start of each route, cutting any upward command to the room left below the cap.
With the cap, their SR is 36.44\%, 45.25\%, and 49.65\%, against 50.15\%, 58.36\%, and 66.37\% without it on the same routes, so $\pi_3$ keeps 13.21 of its 16.22-point gain over $\pi_0$ (Fig.~4(f) of the main paper).

\end{document}